\documentclass[11pt]{article}
\usepackage{winlp2021}
\usepackage{times}
\usepackage{url}
\usepackage{latexsym}

\winlpfinalcopy 

\title{Coral: An Approach for Conversational Agents in Mental Health Applications}

\author{Harsh Sakhrani\textsuperscript{\textsection}, Saloni Parekh\textsuperscript{\textsection} \and Shubham Mahajan\textsuperscript{\textsection} \\
Pune Institute of Computer Technology\\
Pune, India\\
{\tt harshsakhrani26@gmail.com, saloniparekh1609@gmail.com}\\
{\tt svmahajan8899@gmail.com} \\}

\date{}
\begin{document}
\maketitle
\begingroup\renewcommand\thefootnote{\textsection}
\footnotetext{Equal Contributors}
\endgroup
\begin{abstract}
  It may be difficult for some individuals to open up and share their thoughts and feelings in front of a mental health expert. For those who are more at ease with a virtual agent, conversational agents can serve as an intermediate step in the right direction. The conversational agent must therefore be empathetic and able to conduct free-flowing conversations. To this effect, we present an approach for creating a generative empathetic open-domain chatbot that can be used for mental health applications. We leverage large scale pre-training and empathetic conversational data to make the responses more empathetic in nature and a multi-turn dialogue arrangement to maintain context. Our models achieve state-of-the-art results on the Empathetic Dialogues test set.
\end{abstract}

\section{Introduction}
\label{intro}

%
%

\blfootnote{
    %
    %
    %
    %
    \hspace{-0.65cm}  
    This work is licensed under a Creative Commons 
    Attribution 4.0 International Licence.
    Licence details:
    \url{http://creativecommons.org/licenses/by/4.0/}.
     
    %
}

Conversations pertaining to mental health touch upon deep and fragile feelings and deal with one finding solace with words. To create a conversational agent that can initiate such discussions or assist someone in distress, it must be capable of having empathetic conversations. To this end, we propose Coral, a conversational agent designed using the Generative Transformer Decoder \cite{dialo} that incorporates empathy and acts as an intermediate step for people who feel more comfortable with a virtual agent. Another significant issue that Coral can tackle is that, like any human, it considers the context of the conversation while generating its next response. Thus, we make the following contributions:
$(i)$ We propose Coral, a multi-turn open domain chatbot, pre-trained on large scale social media discussion chains. We make use of the Generative Transformer Decoder as our major building block and leverage empathetic conversational data to make it more suitable for mental health applications.
$(ii)$ We empirically show that the model can maintain context in its responses owing to the fact that the model was trained in a multi-turn conversation setting, where the input sequence is all turns of the context and the output sequence is the response.
$(iii)$ Furthermore, our experimental results demonstrate that we outperform the existing models in terms of the Perplexity and Average BLEU score. 

\section{Methodology}
In this section, we first explain the Generative Transformer Decoder (GTD) Block and its pre-training strategy, followed by the multi-turn dialogue arrangement. This is followed by the fine-tuning setup and the hyperparameter details.

{\bf Pre-Training Strategy}:
The generative transformer decoder block is inspired by the DialoGPT \cite{dialo} model architecture. The internal architecture is significantly dominated by the Transformer Decoder \cite{attention}. The model is formulated as an auto-regressive language model but was pre-trained on large-scale dialogue pairs extracted from Reddit discussion chains. This enables the model to capture conversational flow with finer granularity, making it more suitable for conversational applications. 
We utilise two variants of the GTD's publicly available weights: $(i)$. \textbf{12} Transformer Decoders (Coral small)\footnote{https://huggingface.co/microsoft/DialoGPT-small} $(ii)$. \textbf{24} Transformer Decoders (Coral large).\footnote{https://huggingface.co/microsoft/DialoGPT-medium}.

{\bf Multi-Turn Dialogue Arrangement}:
As mentioned earlier, GTD frames the response generation task as a language modelling task. All dialogue turns within a dialogue session are first concatenated into a long text $x_1$ ,..., $x_B$ ($B$ is the sequence length) concluded by the end-of-text token. The source sequence is denoted as
$S$ = $x_1$ ,..., $x_a$, where the source sequence length is $a$, while the target sequence (groundtruth response) is denoted as $T$ = $x_{a+1}$ ,..., $x_B$, where the target sequence length is $B - a + 1$. The conditional probability $p(T|S)$ can be written as the product of a series of conditional probabilities:
$p(T|S) = \prod_{n=a+1}^{B} p( x_n | x_1,...,x_{n-1})$. For a multi-turn dialogue session $T_1$ ,..., $T_H$, where $H$ is the number of dialogues turns, the above equation can be written as $p(T_H$ ,..., $T_2 | T_1)$, which in essence is the product of conditional probabilities of $p(T_i | T_1$ ,..., $T_{i-1})$ where $i$ $\in$ $\{1, H\}$.


{\bf Finetuning for Empathy}:
One of the key requirements for a mental health conversational agent is recognizing feelings in a conversation and replying accordingly. To this end, we make use of the Empathetic Dialogues dataset \cite{empathetic}. The dataset contains 25k open domain one-to-one conversations grounded in emotional situations. We use the Negative Log-Likelihood function to calculate the loss in the fine-tuning phase.

{\bf Hyperparameter Setup}:
The hyperparameter setup is common for all Coral variants. We use the Adam optimizer with the epsilon value $10^{-8}$. The batch size for both training and evaluation was chosen to be 4. The initial learning rate was set to 5 x $10^{-5}$. The model was trained for 3 epochs on the Nvidia TESLA K80 GPU.


\begin{table}[t]
  
  \centering
  \begin{tabular}{l|c|c}
    \hline
    Model &Perplexity &Average BLEU\\
    \hline
    
    Pre-Trained \cite{empathetic}  &27.96 &5.01\\
    TopicPrepend \cite{empathetic} &25.02 &3.13\\
    EmoPrepend \cite{empathetic} &23.96 &2.69\\
    Fine-Tuned \cite{empathetic} &21.24 &6.27\\
    Ensem-DM \cite{empathetic} &19.05 &6.83\\  
    CAiRE \cite{caire} &13.32 &7.03\\
    \textbf{Coral (CW-2, small)} &\textbf{9.12} &\textbf{23.49}\\
   \textbf{Coral (CW-4, small)} &\textbf{7.86} &\textbf{22.12}\\
   \textbf{Coral (CW-6, small)} &\textbf{12.85} &\textbf{22.26}\\
    \textbf{Coral (CW-6, large)\footnotemark}  &\textbf{7.63} &\textbf{22.91}\\
    \hline
  \end{tabular}
  \caption{\label{table:tab2}Comparison of Perplexity and Average BLEU on Empathetic Dialogues Dataset. CW stands for Context Window}
\end{table}

\footnotetext{{We could not perform additional experiments on this variant due to computational constraints.}}
\begin{table}[t]
  
  \centering
  \begin{tabular}{l|l}
    \hline
    \textbf{Coral} &\textbf{Vanilla GTD}\\
    \hline
    \textbf{Conversation 1:} &\textbf{Conversation 1:} \\
    \hline
    \textit{User:}  I feel guilty of cheating on the test &\textit{User:}  I feel guilty of cheating on the test\\
    \textit{Coral:} Why did you cheat on it? &\textit{GTD:} I feel guilty for not being able to read the test.\\
    \textit{User:}  I did not study &\textit{User:}  Why do you feel guilty about that?\\
    \textit{Coral:} I understand. I hope you will find a way &\textit{GTD:} I feel guilty about that. 
    \\to improve your score\\
    \hline
    \textbf{Conversation 2:} &\textbf{Conversation 2:} \\
    \hline
    \textit{User:}  I have had a great start to my week! &\textit{User:}  I have had a great start to my week!\\
    \textit{Coral:} That's great! How have you been doing? &\textit{GTD:} I have a great start to my week!\\
    \textit{User:}  I think I have just been a positive person &\textit{User:}  That's good for you\\
    \textit{Coral:} That's great! I hope you continue to be  &\textit{GTD:} I have had a great start to my week!
    \\a positive person!  \\
    \hline
    
  \end{tabular}
\caption{\label{table:tab1}Examples of interaction with Coral (small) and vanilla GTD.}
\end{table}

\section{Results and Analysis}
For the comparison against previous techniques, we report Perplexity and the Average BLEU score in Table~\ref{table:tab2} for Conversational Response Generation on the Empathetic Dialogues Test Set. Table~\ref{table:tab1} shows the comparison between Coral and vanilla GTD. Our main observations after performing this study have been: $(\textbf{i})$ The model has better conversation capability owing to the wide-scale conversational pre-training and the large number of parameters. This is especially demonstrated by the significantly improved performance on the two evaluation metrics. $(\textbf{ii})$ We observe that, for all Coral variants, the Average BLEU score performance is almost identical, however, there is a considerable variation in the Perplexity performance. This variation can be explained by the fact that each context-window size has an unequal number of training utterances(~69000 training utterances for 2,~13500 training utterances for 4 and ~2500 training utterances for 6). However, despite the fact that window-size 4 has significantly fewer training utterances, a substantial drop in perplexity from window-size 2 to window-size 4 suggests that increasing the context-window size leads to better Perplexity performance. From window-size 4 to window-size 6, however, there is a decrease in the Perplexity performance, which can be attributed to two factors: 1. Lesser training utterances 2. A larger context-window size may be counterproductive. $(\textbf{iii})$ As demonstrated in Table 2, there is a substantial difference between the responses generated by Coral and those generated by vanilla GTD. The fact that Coral was fine-tuned on Empathetic Conversational data explains this distinction.

Another method to evaluate empathy in a given computer-generated conversation would be to incorporate human evaluation like in \cite{dialo}. We faced two problems when we considered this approach. Firstly, human evaluation approaches like \cite{dialo} need a large number of test samples generated by humans and multiple conversational agents for comparison. Given the resources we had for this research project, this was not a possibility. Secondly, as demonstrated in \cite{dialo}, human evaluation may tend to give unexpected results because of inconsistencies in human answers as well as evaluation.

\section{Ethical Considerations}
Certain ethical considerations have been made by \cite{dialo} on the text obtained from Reddit- a social media platform. Because of the nature of social media text, the language model is likely to be affected by obscene and inappropriate language. To prevent this, \cite{dialo} removed potentially offensive language from Reddit conversations by eliminating sub-reddits and instances where the target is likely to include inappropriate language. But this indirectly aggravates the already existing problem of under-representation of the minority groups on social media; by removing words that we believe to be offensive, we may unintentionally remove the discourse of these minority groups that otherwise describes them in a positive light \cite{blender}. From a mental health point of view, EmpatheticDialogues \cite{empathetic} consists of one-on-one conversations which may provide comfort to the user. However, due to the complexity of mental health problems, humans must be extremely careful when they converse with a user. The model is no exception but is more likely to generate a response that could act as a trigger for the user. Additionally, the model is more likely to lack the ability to accurately respond to what the user is looking for.


\end{document}